\documentclass[letterpaper, 10 pt, conference]{IEEEtran}  % Comment this line out if you need a4paper

\IEEEoverridecommandlockouts                              % This command is only needed if 
\usepackage[numbers, sort]{natbib}
\usepackage[utf8]{inputenc} % allow utf-8 input
\usepackage[T1]{fontenc}    % use 8-bit T1 fonts
\usepackage{hyperref}       % hyperlinks
\usepackage{url}            % simple URL typesetting
\usepackage{booktabs}       % professional-quality tables
\usepackage{amsfonts}       % blackboard math symbols
\usepackage{nicefrac}       % compact symbols for 1/2, etc.
\usepackage{microtype}      % microtypography
\usepackage{graphicx}
\usepackage{amsmath,amssymb,amsfonts}
\usepackage[acronym]{glossaries}
\usepackage{comment}
\usepackage{xcolor}
\usepackage{enumerate}
\usepackage[capitalise]{cleveref}
\usepackage{caption}
\usepackage{svg}
\usepackage{listings}
\usepackage{siunitx}
\usepackage[section]{placeins}
\usepackage[english]{babel}
\usepackage{booktabs}
\usepackage{multirow}
\usepackage{cancel}
\usepackage[normalem]{ulem}
\useunder{\uline}{\ul}{}
\usepackage{eso-pic}

\newcommand\AtPageUpperMyleft[1]{\AtPageUpperLeft{%
\put(\LenToUnit{1cm},\LenToUnit{-2cm}){#1}%
}}%

\AddToShipoutPictureBG*{%
  \AtPageUpperMyleft{\parbox[b][2cm][c]{\paperwidth}{%
    \centering
    \fontsize{12}{14}\selectfont
    \color{gray!50}
    This paper has been accepted for publication at the\\
    IEEE International Conference on Robotics and Automation (ICRA), London 2023. \copyright{}IEEE
  }}%
}

\AddToShipoutPictureBG*{
  \AtPageLowerLeft{%
    \raisebox{25pt}{\makebox[\paperwidth]{\begin{minipage}{21cm}\centering
    \fontsize{10}{12}\selectfont
      \textcolor{gray!50}{ \copyright 2023 IEEE.  Personal use of this material is permitted.  Permission from IEEE must be obtained for all other uses, in any current or future media, including reprinting/republishing this material for advertising or promotional purposes, creating new collective works, for resale or redistribution to servers or lists, or reuse of any copyrighted component of this work in other works.
      }
    \end{minipage}}}%
  }
}
\newcommand{\rev}[1]{\textcolor{black}{#1}}
\newcommand{\revdel}[1]{\textcolor{red}{}}
\title{\LARGE \bf \vspace{6mm}
Model- and Acceleration-based Pursuit Controller for High-Performance Autonomous Racing
}
\author{Jonathan Becker, Nadine Imholz, Luca Schwarzenbach \\ Edoardo Ghignone, Nicolas Baumann and Michele Magno % <-this % stops a space
\thanks{Manuscript received 2 September 2022; revised 6 February 2023; accepted 17 January 2023. All authors are associated with Center for Project Based Learning, D-ITET, ETH Zurich.
\emph{(Jonathan Becker, Nadine Imholz, Luca Schwarzenbach, Edoardo Ghignone and Nicolas Baumann contributed equally to this work.) (Corresponding author: Nicolas Baumann.)}}
}

\newacronym{mpc}{MPC}{Model Predictive Control}
\newacronym{mpcc}{MPCC}{Model Predictive Contouring Controller}
\newacronym{rl}{RL}{Reinforcement Learning}
\newacronym{mlp}{MLP}{Multilayer Perceptron}
\newacronym{forl}{FoRL}{Foundations of Reinforcement Learning}
\newacronym{ml}{ML}{Machine Learning}
\newacronym{sb3}{SB3}{Stable Baselines 3}
\newacronym{sac}{SAC}{Soft Actor Critic}
\newacronym{ppo}{PPO}{Proximal Policy Optimization}
\newacronym{ai}{AI}{Artificial Intelligence}
\newacronym{nn}{NN}{Neural Network}
\newacronym{sota}{SotA}{State-of-the-Art}
\newacronym{esc}{ESC}{Electronic Speed Controller}
\newacronym{ros}{ROS}{Robot Operating System}
\newacronym{imu}{IMU}{Inertial Measurement Unit}
\newacronym{ekf}{EKF}{Extended Kalman Filter}
\newacronym{slam}{SLAM}{Simultaneous Localization And Mapping}
\newacronym{sdc}{SDC}{Self Driving Cars}
\newacronym{obc}{OBC}{On Board Computer}
\newacronym{qp}{QP}{Quadratic Programming}
\newacronym{uav}{UAV}{Unmanned Aerial Vehicles}
\newacronym{cg}{CG}{Center of Gravity}
\newacronym{em}{EM}{Expectation Maximization}
\newacronym{rms}{RMS}{Root Mean Square}
\newacronym{map}{MAP}{Model- and Acceleration-based Pursuit}
\newacronym{pd}{PD}{Proportional-Derivative}
\newacronym{lut}{LUT}{Lookup Table}
\begin{document}

\pdfcompresslevel=9

\maketitle
\thispagestyle{empty}
\pagestyle{empty}

%%%%%%%%%%%%%%%%%%%%%%%%%%%%%%%%%%%%%%%%%%%%%%%%%%%%%%%%%%%%%%%%%%%%%%%%%%%%%%%%
\begin{abstract}
Autonomous racing is a research field gaining large popularity, as it pushes autonomous driving algorithms to their limits and serves as a catalyst for general autonomous driving. For scaled autonomous racing platforms, the computational constraint and complexity often limit the use of \gls{mpc}. As a consequence, geometric controllers are the most frequently deployed controllers. They prove to be performant while yielding implementation and operational simplicity. Yet, they inherently lack the incorporation of model dynamics, thus limiting the race car to a velocity domain where tire slip can be neglected. This paper presents \gls{map} a high-performance model-based trajectory tracking \rev{controller} that preserves the simplicity of geometric approaches while leveraging tire dynamics. The proposed algorithm allows accurate tracking of a trajectory at unprecedented velocities compared to \gls{sota} geometric controllers. The \gls{map} controller is experimentally validated and outperforms the reference geometric controller four-fold in terms of lateral tracking error, yielding a tracking error of \SI{0.055}{m} at tested speeds up to \SI{11}{\meter/\second} \rev{on a scaled racecar}. \rev{Code: \url{https://github.com/ETH-PBL/MAP-Controller}.}

\end{abstract}

%%%%%%%%%%%%%%%%%%%%%%%%%%%%%%%%%%%%%%%%%%%%%%%%%%%%%%%%%%%%%%%%%%%%%%%%%%%%%%%%
\section{INTRODUCTION}

Motorsport racing has proven to enable knowledge transfer of cutting-edge research to the automotive industry \cite{catalyst0, catalyst1, ar_survey}. In particular, autonomous racing presents a new frontier that promises to revolutionize autonomous driving by enabling and stress-testing new technologies and algorithms in the field of \gls{sdc} \cite{dronerace_jfr, f110, indyautonomous, roborace}. For this reason, many autonomous racing competitions have recently emerged, featuring different platforms and form factors, from full-scaled Indy Autonomous \cite{indyautonomous} and Formula Student Driverless \cite{amz_fullstack} to scaled F1TENTH \cite{f110, okelly2020f1tenth}. A major challenge in these competitions is the completion of a racetrack in a high-speed setting \cite{ar_survey, indyautonomous, f110, amz_fullstack}, which presents a demanding task for autonomous control systems, as the car's behavior is highly non-linear \cite{mpccurv, liniger_mpcc, story_of_modelmismatch, pacejka1992magic}. This is particularly challenging when the controller needs to operate at the edge of traction, beyond the linear characteristics of the complex tire dynamics \cite{mpccurv, liniger_mpcc, story_of_modelmismatch, pacejka1992magic}. Therefore, tracking trajectories accurately at cornering speeds that saturate the tire friction capacity, presents a challenging and central controls task for autonomous racing \cite{amz_fullstack, indyautonomous, ar_survey}.

Classically, the control strategies in the autonomous racing setting could be loosely categorized into the following three methods, listed in the order of increasing complexity and performance:

\begin{enumerate}[I]
    \item \textbf{Reactive Methods}: Compute control actions directly based on the sensor input they receive \cite{ftg, disparity}. This allows for simple and reactive control, without requiring proper state estimation. However, disregarding information about the track and the car, limits the performance of the controllers, especially in high-speed settings.
    
    \item \textbf{Geometric Methods}: Utilize the geometric properties of the vehicle, typically derived from the kinematic single-track model \cite{bicycle_model}. Methods such as \emph{Pure Pursuit} \cite{pp} or \emph{Stanley Control} \cite{stanley_jfr} separate longitudinal and lateral control. While these methods yield a great racing performance advantage over their reactive counterparts, they do not take into account the dynamic model of the vehicle, thus leading to sub-optimality.
    
    \item \textbf{Model-Based Controllers}: A typical example of a model-based controller is the family of \gls{mpc} methods. They consist of applying optimal control over a receding horizon and can take into account the dynamic model of the vehicle as well as state and input constraints thus yielding highly performant autonomous racing results \cite{liniger_mpcc, mpccurv, amz_fullstack}. However, \rev{precise state-estimation and knowledge of the car model and delays present in the system are essential to achieve good \gls{mpc} performance, making this method the most computationally demanding and difficult to implement control strategy of the aforementioned methods \cite{GONZALEZ2019213}.} \revdel{This method represents by far the most complex and computationally demanding control strategy among the aforementioned methods. Therefore simpler methods are often preferred \cite{fuchs2021, songgts}.}
\end{enumerate}

\revdel{
More recently, many emerging applications of small autonomous robots are being deployed with strict energy constraints. For instance, the \emph{Pico Drone} is a small/battery-operated robot that is thermally restrained and enhances a small enclosed fan-less form factor, which significantly limits the computational resources as it can only operate with embedded processors \cite{duisterhof2021tiny, an2021real}. Constrained hardware introduces unforeseen challenges in implementing control capability in high-speed application scenarios such as autonomous nano-drones or autonomous racing where latency and real-time throughput need to be guaranteed \cite{roborace, liniger_mpcc}.
}

\rev{
The development of autonomous race cars brings about many often unforeseen challenges. Especially at small scale, energy and size constraints necessitate the deployment on embedded processors \cite{duisterhof2021tiny, an2021real} with limited computational resources. Constraints on sensors make state-estimation more difficult compared to full-scale cars and often introduce delays. Yet, for high-speed applications low latency and real-time throughput need to be guaranteed \cite{liniger_mpcc}. Many refrain from deploying \gls{mpc} due to the added implementational and computational complexity given these constraints and prefer simpler alternatives \cite{pp, ar_survey}. In order to maintain the performance advantage of model-based control, we propose a way of incorporating model knowledge into the simple control architecture of geometric controllers.
}

This paper introduces \gls{map}, a high accuracy and low complexity model-based lateral controller to track an arbitrary raceline trajectory in a high-speed autonomous racing scenario. The controller \revdel{is designed for accurate trajectory tracking in a high-speed autonomous racing setting and} has been evaluated on the 1:10 scaled F1TENTH autonomous racing resource-constrained platform \cite{f110, okelly2020f1tenth}. Yet, it is to be mentioned, that the control strategy can be applied to full-scaled vehicles as well.  
 
 The proposed \gls{map} controller incorporates the highly non-linear \emph{Pacejka} tire model \cite{pacejka1992magic} into lateral geometric control, inspired by aerospace fixed-wing $L_1$ guidance \cite{l1}. This results in a more accurate control strategy that would need to be newly classified between geometric and model-based methods, as it leverages the accuracy of model-based lateral controls with the simplicity of a geometric controller. Furthermore, the proposed controller is computationally efficient enabling real-time operation on the computationally constrained \gls{obc} platform. The controller has been experimentally evaluated on the F1TENTH testing platform while racing at a tested top speed of \SI{11}{\meter/\second}. The main contribution of the paper is the design and evaluation of the proposed \gls{map} controller which offers the following benefits:

\begin{itemize}
    \item \textbf{High-Performance:} The model-based lateral controller results in a four-fold higher tracking accuracy at high velocity compared to geometric controllers such as \emph{Pure Pursuit}. Namely, \SI{0.055}{m} lateral error at speeds up to \SI{11}{\meter/\second}. By incorporating the tire dynamics, the controller is able to \rev{compute} the steering angle necessary to track the desired trajectory with great accuracy.
    
    \item \textbf{Low Complexity:} As the controller still separates longitudinal and lateral control, it is simple to implement, comparable to a geometric controller, yet has the benefit of high accuracy by including model-based properties. Thus, it is significantly less complex to implement and optimize than a full-fledged \gls{mpc}.
    
    \item \textbf{Computational Efficiency:} The computation of a single control cycle amounts to \SI{6}{ms} \footnote{On a single embedded ARMv8 core of the NVIDIA Jetson NX CPU}. Therefore the low latency of the controller allows for real-time operation in a high-speed setting, on computationally constrained hardware. 
\end{itemize}

\section{RELATED WORK}
\begin{comment}
The current \gls{sota} high-performance racing controllers are based on optimal control methods such as \gls{mpc} \cite{amz_fullstack, jain_bayesrace_2020, frohlich2021model}. \gls{mpc} can offer the ability to optimally plan and track a racing trajectory, even incorporating system and safety constraints, yet at the expense of complexity and computation, \cite{fuchs2021, songgts} of formulating the \gls{qp} problem. Thus it is time-consuming and often non-feasible to tune the \gls{mpc}, to achieve a performant controller \cite{story_of_modelmismatch}, requiring a domain expert. 

However, for the reason of complexity and computational burden, \gls{mpc}
\end{comment}

The current \gls{sota} high-performance racing controllers, on full-scaled vehicles, are based on optimal control methods such as \gls{mpc} \cite{amz_fullstack, jain_bayesrace_2020, frohlich2021model}. \gls{mpc} can offer the ability to optimally plan and track a racing trajectory, even incorporating system and safety constraints, yet at the expense of complexity and computation, \cite{fuchs2021, songgts} of formulating and solving the \gls{qp} problem. %Thus it can be extremely difficult to tune the \gls{mpc}, to achieve a performant controller, requiring a domain expert \cite{story_of_modelmismatch}.

However, for the reason of complexity and computational burden, \gls{mpc} has rarely been used in scaled and computationally constrained autonomous racing competitions \cite{ar_survey} such as F1TENTH or drone racing \cite{song_drones_2021}, where the robotic vehicles need to work with limited processing power. High-performance racing requires upper latency bounds of a few tens of milliseconds \cite{liniger_mpcc}, therefore onboard processing on low-power embedded computing platforms calls for a computationally lightweight controller \cite{fuchs2021, songgts}.

 To overcome the computational limits, the most utilized high-performance controller in scaled autonomous racing is of the geometric type, most commonly a \emph{Pure Pursuit} controller \cite{pp}. The \emph{Pure Pursuit} controller is a lateral controller, that pursues a point on the raceline that lies in front of the vehicle, the distance towards said point is defined as the lookahead distance $L_d$, from which one can compute the necessary steering angle based on the \emph{Ackermann} steering model \cite{althoff_commonroad_2017}. As $L_d$ is its only tune-able parameter, setting the $L_d$ correctly is crucial for high-performance racing. In the traditional \emph{Pure Pursuit} method \cite{pp}, the $L_d$ is a chosen constant. This inherently presents a trade-off; a too-short $L_d$ produces oscillations at high velocities; too high $L_d$ results in inaccurate tracking and the cutting of corners. 

While the \emph{Pure Pursuit} controller plays an important role in the domain of \gls{sdc} and autonomous racing, the downsides of this control strategy become apparent once the controller is pushed to the physical edge of traction \cite{liniger_mpcc, mpccurv}. 
\rev{Under high lateral acceleration} the car's driving behavior becomes highly non-linear, which pure geometric controllers fail to capture, resulting in tracking errors \cite{liniger_mpcc, mpccurv}. 
%This makes the \emph{Pure Pursuit} controller scale badly with velocity. To address this problem, the $L_d$ needs to be scaled with the velocity. 
Previous works proposed methods of dynamically changing the lookahead distance $L_d$ of Pure Pursuit, to mitigate this effect \cite{adapt_pp0, adapt_pp1, adapt_pp2}. Others introduced fuzzy logic based on the lateral deviation and track curvature \cite{fuzzy_pp0, fuzzy_pp1, fuzzy_pp2}, they however did not address the underlying issue of disregarding tire slippage.

Similar to \emph{Pure Pursuit}, a lateral guidance law can be found in the aerospace domain for fixed-wing \gls{uav}s, namely the $L_1$ guidance \cite{l1}. This nonlinear guidance logic is designed for curved trajectory tracking of a fixed-wing \gls{uav}, by choosing a reference point on the trajectory in front of the aircraft, namely the $L_1$ distance. Therefore, the similarities to \emph{Pure Pursuit} with its lookahead distance $L_d$, can be seen. However, \emph{Pure Pursuit} commands a steering angle, whereas as the $L_1$ guidance outputs a lateral acceleration $a_{lat}$ governed by \cref{eq:l1}, where $V$ is the velocity and $\eta$ is the angle between the velocity vector and the reference point.

\begin{equation}
    a_{lat} = 2\frac{V^2}{L_1}sin(\eta)
    \label{eq:l1}
\end{equation}

This paper proposes and demonstrates how to leverage the acceleration-based guidance signal $a_{lat}$ to incorporate model knowledge in a geometric control setting while applying well-known adaptive $L_d$ techniques. The \emph{Pacejka Magic Tire Formula} \cite{pacejka1992magic} allows the computation of the required $a_{lat}$ to map to the steering angle. The incorporation of the dynamic model allows the controller to consider tire slip while inherently scaling with the velocity, allowing for accurate lateral control in the high-speed and non-linear domain. This yields a high-performance racing controller that outperforms \gls{sota} geometric racing controllers in terms of lap time and four times in tracking error. The controller has been fully designed, implemented, and evaluated in a scaled F1TENTH car \cite{F1TENTHbuild}. 

\section{METHODOLOGY}
In this section, we show through abstraction that \emph{Pure Pursuit} uses the same $L_1$ guidance law that was proposed for \gls{uav}s \cite{l1}. With this, we show that \emph{Pure Pursuit} disregards wheel slip, an assumption that fails at high speeds. We propose a method able to better capture and control the real-world dynamics of the car. This is done by identifying the car dynamics in \cref{subsec:tires} and using them to convert guidance to control commands via a \gls{lut} described in \cref{subsec:lut}. Finally, in \cref{subsec:adaptive} a mathematical reason for changing the lookahead distance is given.

\subsection{Proposed Control System Overview} \label{subsec:cs_overview}
The $L_1$ guidance of \cite{l1} results in a desired centripetal acceleration $a_c$ of the vehicle, equal to the tangential speed $v_t$ multiplied with the yaw rate $\dot{\Psi}$:
\begin{equation}
    a_{c} = v_t \cdot \dot{\Psi} = 2 \frac{v_t^2}{L_d}sin(\eta)
    \label{eq:l1_guidance}
\end{equation}
Where $L_d$ is the lookahead distance and $\eta$ is the angle between the velocity vector and lookahead point. 

\begin{figure}
    \centering
    \includegraphics[width=\columnwidth]{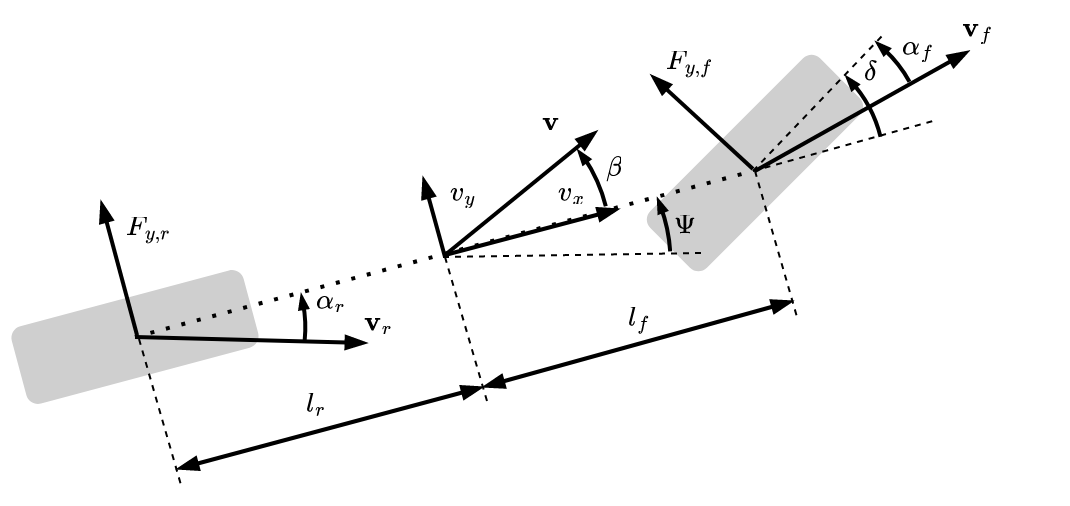}
    \caption{Overview of the single-track model \cite{althoff_commonroad_2017} with the relevant forces and angles defined in their positive directions.}
    \label{fig:single_track}
\end{figure}

\emph{Pure Pursuit} can be derived from this equation, by mapping the lateral acceleration to a steering angle under the assumption of \emph{Ackermann} steering. This assumes that no side slip occurs and the longitudinal velocity in the car's frame $v_x = v_t$:
\begin{align}
    \delta &= \tan^{-1}\left( \frac{a_{c} \cdot l_{wb}}{v_x^2}\right)
\end{align}
where $l_{wb}$ is the length of the wheelbase. Inserting the desired acceleration from \cref{eq:l1_guidance} yields the formula used in \emph{Pure Pursuit}:
\begin{equation}
    \delta = \tan^{-1}\left( \frac{2 sin(\eta) l_{wb}}{L_d}\right)
\end{equation}
When driving corners at higher speeds, tire slip starts to occur and the underlying assumption of \emph{Ackermann} steering from \emph{Pure Pursuit} no longer holds. The desired acceleration $a_c$ is no longer correctly commanded and the car starts to drift away from the trajectory. Therefore, we propose a method of incorporating the tire slip into the conversion from centripetal acceleration to the steering angle. This is done, by formulating the single-track model \cite{bicycle_model} in terms of the lateral forces acting on the tire, which yields the following equations of motion for the lateral velocity $v_y$ and the yaw-rate $\dot{\Psi}$, with forces and angles defined in \cref{fig:single_track}.
\begin{align}
    \dot{v}_y &= \frac{1}{m}\left[ F_{y,r} + F_{y,f} \right] - v_x \cdot \dot{\Psi} \label{eq:v_y}\\
    \ddot{\Psi} &= \frac{1}{I_z} \left[ -l_r F_{y,r} + l_f F_{y,f}\right] \label{eq:psiddot}
\end{align}
Where $m$ is the mass of the vehicle and $I_z$ is the moment of inertia around the yaw axis. The lateral forces produced by the front (index $f$) and rear axle (index $r$) can be related to the tire slip angles $\alpha$ with the general formula ($i \in (f, r)$):
\begin{equation}
    F_{y,i} = \mu F_{z,i} \cdot f_i(\alpha_i)
    \label{eq:lateral_forces}
\end{equation}
where $\mu$ is a friction coefficient and $F_{z,i}$ the vertical load on axle $i$. The respective tire slip angles $\alpha$ are found through the relation:
\begin{align}
    \alpha_f &= \arctan\frac{v_y + \dot{\Psi} l_f}{v_x} - \delta \\
    \alpha_r &= \arctan\frac{v_y - \dot{\Psi} l_r}{v_x} 
\end{align}
And the load on the front and rear axle are found depending on the longitudinal acceleration $a_{long}$ and the mass distribution of the car, depending on the height of the \gls{cg} above the ground $h_{cg}$:
\begin{align}
    F_{z,f} &= \frac{m g l_r - m a_{long} h_{cg}}{l_r + l_f} \\
    F_{z,r} &= \frac{m g l_f + m a_{long} h_{cg}}{l_r + l_f}
\end{align}
The steering angle $\delta$ enters these equations only by affecting the front tire slip angle $\alpha_f$. Hence, it is necessary to find a model for the lateral force before being able to map the desired acceleration to a steering angle. 
\subsection{Tire Model} \label{subsec:tires}
\begin{figure}
    \centering
    \includegraphics[width=\columnwidth]{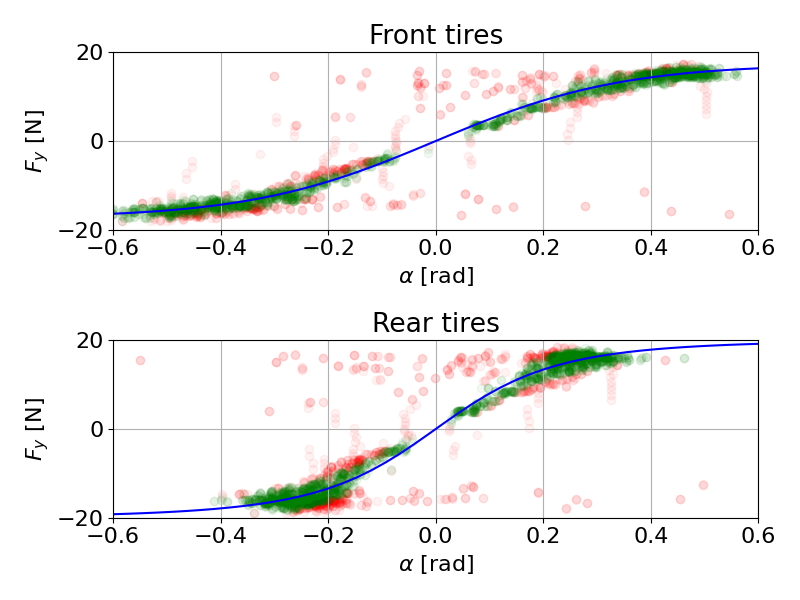}
    \caption{Data points obtained from the steady-state cornering experiment and the resulting model fit of the Pacejka model. Outliers are marked red, inliers green, and the model prediction is shown in blue for a fixed load of 16.4\,N on the front axle and 18.6\,N on the rear axle.}
    \label{fig:pacejka_fit}
\end{figure}

The \emph{Pacejka Magic Formula} model \cite{pacejka1992magic} is able to accurately capture tire slip and model the transition of static to dynamic friction. In the original formulation, the authors included shift factors that are omitted in our version for simplicity. The model relates the tire slip angle at axle $i$ to the respective lateral force  $F_{y,i}$ via the non-linear relationship:
\begin{equation}
\begin{aligned}
    F_{y,i} = {} & \mu F_{z,i} \cdot D_i \sin [ C_i \cdot \\
    & \arctan \left(B_i \alpha_i - E_i \left( B_i \alpha_i - \arctan (B_i \alpha_i)\right)\right)]
    \label{eq:magic}
\end{aligned}
\end{equation}
The authors described the parameters as follows:
\begin{itemize}
    \item $B$ : Stiffness Factor
    \item $C$ : Shape Factor
    \item $\tilde{D}$ : Peak Value
    \item $E$ : Curvature Factor
\end{itemize}
Where in \cref{eq:magic} the peak value $\tilde{D}$ is in fact $\mu F_{z,i} \cdot D_i$. 
A data-driven method was employed to identify these parameters. The data was gathered in steady-state cornering experiments as proposed by Voser et. al. \cite{sysidVoser}, in which the car was driven at a constant speed and the steering angle was increased slowly (0.02\,rad/s). This made it possible to relate the lateral forces at the front and rear tires to the measured steady-state acceleration by the \gls{imu} $a_{y,imu}$ via the relationship
\begin{align}
    F_{y,f} &= \frac{m \cdot l_r}{ (l_r+l_f) \cdot \cos{\delta}} \cdot a_{y,imu} \\
    F_{y,r} &= \frac{m \cdot l_f}{l_r + l_f} \cdot a_{y,imu}
\end{align}
The parameters were then found through a least squares regression, in which the shape factor $C$ was limited to 1.5 and the curvature factor $E$ was limited to 1.1 to prevent the function to curve back towards zero for high tire slip angles. With this, the data fit in \cref{fig:pacejka_fit} was obtained. Outliers were rejected by performing three steps $k$ of \gls{em} in which only points with an error smaller than $\frac{10}{2^k}$ were kept. This resulted in 25.5\,\% of points being rejected as outliers and a resulting fit with an average residual of 0.87\,N for the front and 1.35\,N for the rear tires.

\subsection{Lookup Table Generation} \label{subsec:lut}
\rev{For a set of fixed steering angles at a certain speed $v_x$, the dynamics of $v_y$ and $\dot{\Psi}$ in \cref{eq:psiddot} and \cref{eq:v_y} converge to a steady-state, where the car turns at a constant radius and with constant centripetal acceleration. Since the guidance law results in smooth changes of commanded centripetal acceleration, the car deviates slightly from this steady-state condition. Therefore, the proposed method is able to neglect the transient phase, commanding steering angles resulting in the desired steady-state centripetal acceleration. The feedback loop and intrinsic stability of the guidance law \cite{l1} mitigate the inaccuracies during the transient phase. %Opposed to that \gls{mpc} simulates the car dynamics online to predict the transient dynamics of these states and incorporates them into the control input. However, these transient dynamics are hard to estimate correctly as they depend strongly on system delays.
}
Since \cref{eq:v_y} cannot be solved analytically for the steady-state centripetal acceleration, a \gls{lut} was generated to obtain a mapping from steering angle to acceleration. For this, the system was simulated using the single-track dynamic model. The state was propagated with a range of constant longitudinal velocities and steering angles for which the resulting steady-state centripetal acceleration was recorded. With this, the controller is able to retrieve the required steering angle for a certain velocity and desired acceleration by interpolating between the closest elements in the table. 

\begin{figure}
    \centering
    \includegraphics[width=\columnwidth]{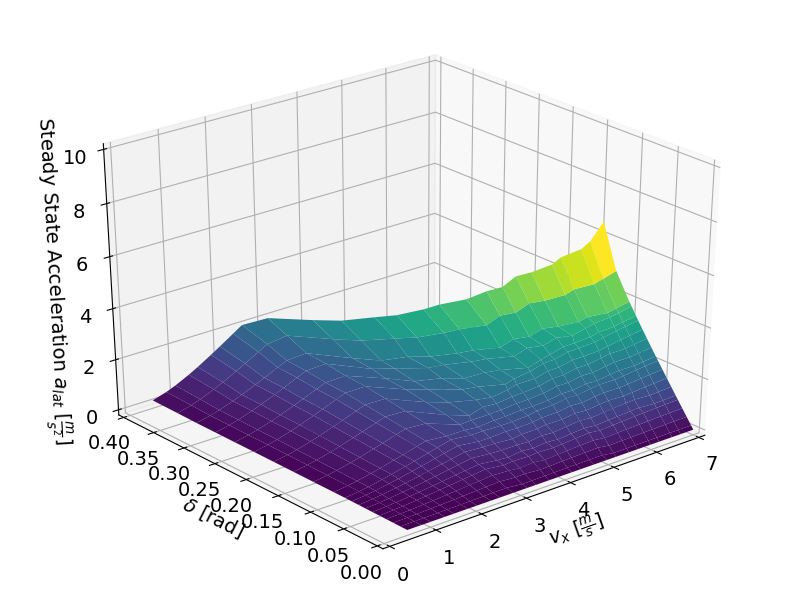}
    \caption{Visualization of the lookup table generated from the identified car model. The steady-state centripetal acceleration is shown as a function of the steering angle and velocity.}
    \label{fig:pacejka_lu}
\end{figure}

\revdel{This approach neglects the transient dynamics of $\dot{v}_y$ in \cref{eq:v_y}, and only considers the resulting steady-state lateral acceleration which comes from the term $v_x \cdot \ddot{\Psi}$, the centripetal acceleration. 
}
As can be seen in the visualization in \cref{fig:pacejka_lu}, there is a lack of solutions for high $\delta$ and $v_x$. In this region, the simulated model did not converge to a steady state acceleration but rather an unstable drift, resulting in an upper bound for the achievable lateral acceleration at a given speed $v_x$.

From an implementation point of view, first, the lateral acceleration $a_{lat}$ for the desired $L_d$ is computed; secondly, the \gls{lut} returns the steering angle based on the model dynamics; from here on the \gls{map} can be treated in the same way as a regular \emph{Pure Pursuit} controller.

\subsection{Adaptive Lookahead Point} \label{subsec:adaptive}
Park et al. showed that the evolution of the lateral distance to the trajectory as a result of the guidance law in \cref{eq:l1_guidance} could be approximated by a second order system with a time constant $\tau=L_d$/$v_x$ and natural frequency $\omega_n = \sqrt{2}v_x/L_d$ \cite{l1}. Therefore, to converge to the trajectory as quickly as possible, $L_d$ should be chosen as small as possible. On the other hand, the natural frequency increases with the velocity. For stability, the natural frequency of the controller must be smaller than half the frequency of the entire vehicle dynamics including delays. Tests showed that for higher speeds this criterion was no longer met and the system became unstable. To address this, $L_d$ was scaled with the velocity \rev{with the affine mapping $L_d = m + qv_{ctrl}$, where $v_{ctrl}$ is the speed commanded to the car, and $m,\,q$ are tunable coefficients. Relating the lookahead distance with speed}  made the natural frequency of the guidance logic independent of the velocity and ensured stability at higher speeds.

% Scaling the L_d distance with velocity: The L1 guidance states that the L1 distance must be chosen larger than the minimum turning distance of the vehicle, as it otherwise becomes unstable. Pure Pursuit and our controller incorporate this, by increasing the distance at higher speeds. It should, however, be noted that for a turn with a larger radius than L1, this scaling has no effect as the acceleration to track that circle remains the same. Essentially, if the steering angle itself is not scaled with the velocity, Pure Pursuit has no way of incorporating tire slip.
%

\section{EXPERIMENTAL RESULTS}
This section presents the lap time and tracking benefits of incorporating the tire dynamics in a high-speed racing setting, as we compare the proposed \gls{map} controller with a purely geometric \emph{Pure Pursuit} controller. Both \emph{Pure Pursuit} and the \gls{map} controller were set up and tuned at equal conditions such that a fair comparison based on lap time and lap-wise \gls{rms} lateral error to the global trajectory can be conducted using an F1TENTH compliant car.

\subsection{Experimental Setup}
The experiments were conducted on an F1TENTH racecar %as in \cref{fig:nuc4}
based on \cite{F1TENTHbuild}, using the \gls{ros}.
The scaled RC car chassis Traxxas Slash 1:10 4WD RTR was equipped with a Hokuyo UST-10LX LiDAR. %and an Intel Core i3-10110U CPU. 
\emph{Pure Pursuit} and the \gls{map} controller were tuned to minimize the lap-wise \gls{rms} lateral error on a reference track while tracking the same global trajectory. The global trajectory was planned with the global trajectory optimizer from \cite{Heilmeier2020MinimumCar}, which generates a minimum curvature path and calculates a velocity profile with a forward-backward solver. %citation needed for forward-backward solver? % by using minimum curvature optimization for the path planning and a forward-backward solver for the velocity profile generation. 
As the velocity profile is computed for perfect tracking, the velocity was scaled down linearly to 60\% for tuning, where 100\% represents \SI{8.5}{\meter/\second} which is a high-speed for scaled tracks and is further highly track-dependent. \rev{The tuning regarded the lookahead distance $L_d$ by varying the aforementioned coefficients $m,\,q$ in \cref{subsec:adaptive} such that the respective controllers minimized the \gls{rms} lateral error to the reference trajectory at a 60\% speed scaling. Allowing both controllers to be tuned individually to their optimal working point.}

\begin{comment}
\begin{figure}[!h]
    \centering
    \includegraphics[width=0.80\columnwidth]{imgs/NUC4.png}
    \caption{The 1:10 scaled RC car chassis Traxxas Slash 1:10 4WD RTR with a Hokuyo UST-10LX LiDAR.} %with the Intel Core i3-10110U CPU as \gls{obc}.}
    \label{fig:nuc4}
\end{figure}
\end{comment}

\subsection{Velocity Scaling of Performance}\label{subsec:speed_scale}
\begin{figure}[!h]
    \centering
    \includegraphics[width=\columnwidth]{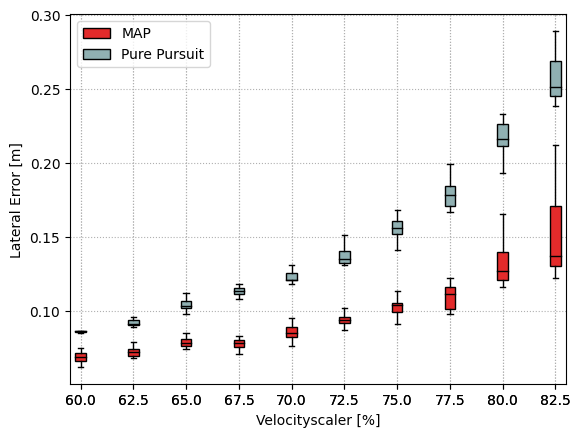}
    \caption{Lap-wise \gls{rms} lateral errors of \emph{Pure Pursuit} and \gls{map} with increasing velocity on the reference track. \rev{The box represents the lower and upper quartile; the whiskers represent the minimum and maximum measured values; the horizontal line represents the median value.} It can be seen that \emph{Pure Pursuit} displays consistently a higher (worse) lateral error with increasing velocity.}
    \label{fig:reftrack}
\end{figure}

To test the performance of both \gls{map} and \emph{Pure Pursuit} at speeds higher than that of tuning, the controllers were tested at different speeds, starting with a velocity scale of 60\% and increasing it in steps of 2.5\% until the controller was no longer capable of maintaining sufficiently accurate tracking, resulting in a crash. For each velocity and controller, 10 rounds were driven with the car. The track used was the same as the one used for tuning.
As can be seen in \cref{fig:reftrack}, lateral error for the \gls{map} controller is significantly lower than that of \emph{Pure Pursuit} at all speeds.
Specifically, at 82.5\% speed, the tracking performance is nearly twice as good for the \gls{map} with respect to the \emph{Pure Pursuit} controller.
In practice, this resulted in \emph{Pure Pursuit} being at the edge of collision with the track boundaries, while \gls{map} was still tracking the trajectory reasonably well.
Regarding lap times, it can be seen that having more accurate trajectory tracking allows for planning more aggressive trajectories ultimately yielding better performance. The \gls{map} controller achieves better lap times at every tested speed, achieving an average improvement of \SI{0.227}{s}. The biggest improvement is  seen at the highest speed, where \gls{map} finishes the lap \SI{0.350}{s} faster than the \emph{Pure Pursuit} controller, corresponding to a $5\%$ time improvement. However, the lower tracking error of the \gls{map} permits tracking even faster trajectories, enabling even higher lap time improvements.

\begin{figure}[!h]
    \centering
    \includegraphics[width=\columnwidth]{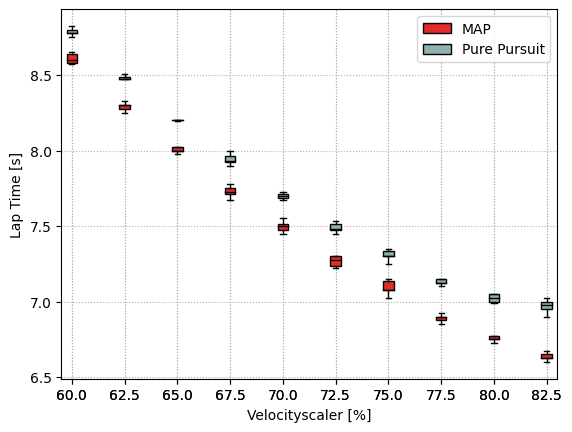}
    \caption{Lap-wise lap times of \emph{Pure Pursuit} and \gls{map}  with increasing velocity on the reference track. \rev{The box represents the lower and upper quartile; the whiskers represent the minimum and maximum measured values; the horizontal line represents the median value.} It can be seen that \emph{Pure Pursuit} consistently displays higher (worse) lap times than the \gls{map}.}
    \label{fig:lapt_vel}
\end{figure}

\begin{comment}
\begin{figure}[!h]
    \centering
    \includegraphics[width=1\columnwidth]{imgs/compare_traj.png}
    \caption{Generalization runs of 10 laps on the \emph{Smile} and \emph{F} track with \gls{map} on the left and \emph{Pure Pursuit} on the right. The velocity was scaled to 75\% on the \emph{Smile} track and to 80\% on the \emph{F} track.}
    \label{fig:traj_compare}
\end{figure}
\end{comment}

\subsection{Effect of Tire Dynamics}
To assess the effect of using different tire models, nominal \gls{map} and \emph{Pure Pursuit} are deployed alongside \gls{map} using a linear tire model\rev{, whose coefficients, that linearly relate force to tire slip angle, are found by using regression on the same data as in \cref{fig:pacejka_fit}}. Tests are conducted at two different reference speed scales, $70\%$ and $80\%$ over five consecutive laps. The resulting lap times are slightly different from the previous paragraph as the track needed to be slightly changed for logistical reasons. 
\begin{figure*}[!h]
    \centering
    \includegraphics[width=\textwidth]{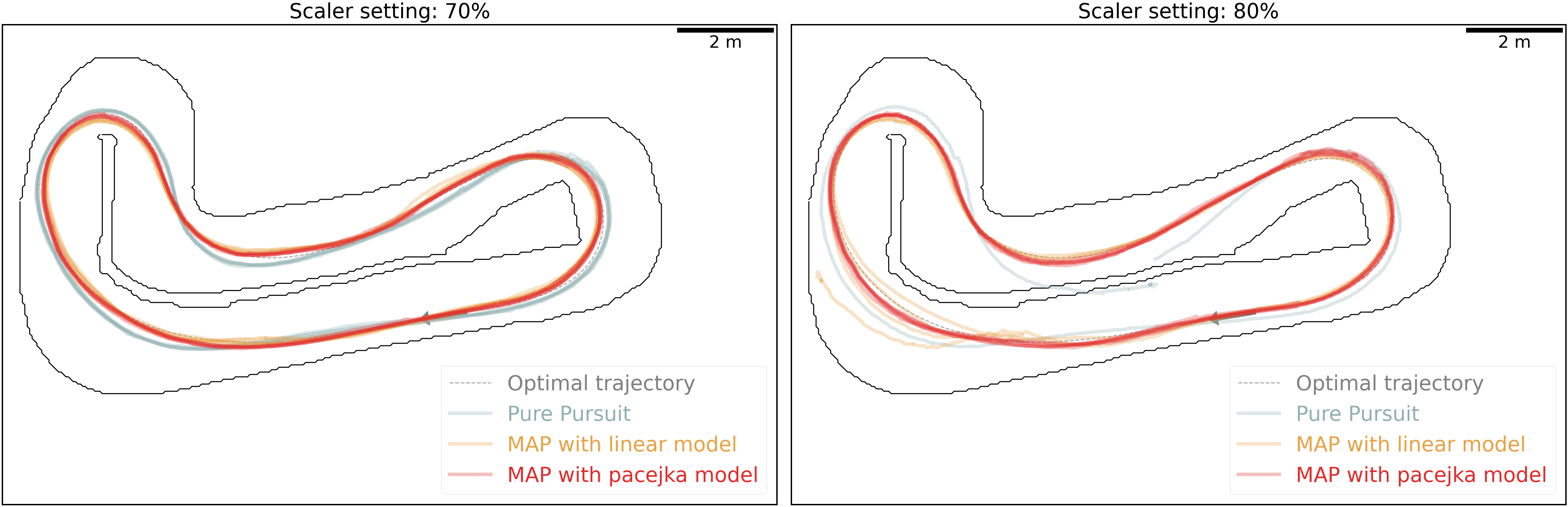}
    \caption{Trajectories of the \emph{Pure Pursuit} controller and the \gls{map} controller with both linear and \emph{Pacejka} tire model. The reference line is shown dashed in gray and an arrow indicates the direction of racing. 
    On the left, the reference speed is scaled to $70\%$ of the nominal one, and to the right to $80\%$. The tracking benefit of accurate tire dynamics can be seen by noticing that in the right plot, at higher speeds, both \emph{Pure Pursuit} and the \gls{map} with linear tire model crash into the track boundaries (driven paths end abruptly at the edge). The \gls{map} controller with the \emph{Pacejka} tire model, instead, manages to successfully track the given trajectory over five laps.}
    \label{fig:pacejkavslinear}
\end{figure*}
In \cref{fig:pacejkavslinear} one can see the different trajectories driven by the two algorithms. The left plots show the results for the slower velocity: at this speed, the controller with the linear tire model is seen cutting the corners. The data in \cref{tab:tire_res} shows that this results in a 2.3\,\% faster lap time at the cost of a 54.2\,\% larger lateral tracking error compared to nominal \gls{map} with the \emph{Pacejka} tire model. \emph{Pure Pursuit} drives wide in all corners. Hence, the nominal \gls{map} outperforms \emph{Pure Pursuit} with 58.2\,\% lower average deviation, 45.5\,\% lower maximum deviation, and 5.1\,\% faster lap time at this speed.

At higher speeds, the benefit of using a more accurate tire model in the controller becomes even more apparent, as the \emph{Pacejka} version of \gls{map} is the only one able to complete five laps, achieving robust tracking of the trajectory. 
The linear version fails to complete a lap at higher speeds, as oscillations cause it to collide with the track boundaries, and \emph{Pure Pursuit} crashes due to under-steering in the third corner. 
The \gls{map} controller with \emph{Pacejka} tire model again achieves the lowest tracking error and deviation. Using the linear tire model resulted in a ten-fold increase in average lateral error during the two laps driven before the crash.
%The resulting improvement in lap time of $~8.8\%$ is only slightly lower than the improvement of the global speed to track ($+14\%$).
It is therefore clear that having a precise tire model capable of capturing the non-linear tire dynamics is key to achieving high performance in autonomous racing.

%Incorporating accurate tire dynamics allows the \gls{map} to perform better in terms of lateral tracking error and lap times, meaning that the proposed controller allows for accurate control at significantly higher velocities, by incorporating lateral tire slip through the tire dynamics. In \cref{fig:pacejkavslinear} we compare the racing  performance of \emph{Pure Pursuit}, \gls{map} with the simplistic linear tire dynamics \cite{althoff_commonroad_2017} and \gls{map} with the more accurate \emph{Pacejka} tire dynamics \cite{pacejka1992magic}. From this experiment and \todo{cref the table}, it is visible that both \gls{map} controllers allow for significantly greater velocities and lower tracking errors than the geometric \emph{Pure Pursuit} counterpart. Between the linear and the more complex and accurate \emph{Pacejka} tire models, it is once again visible, that the inclusion of the more complex non-linear model dynamics allows for higher controller performance at higher velocities. 
%Thus, the benefit of the \gls{map} controller leveraging the \emph{Pacejka} tire dynamics is evident, as it captures the model behavior under lateral tire slip more accurately than the linear tire model and especially the geometric \emph{Ackermann} assumptions which do clearly break down under the evaluated velocity.
\begin{table}[!h]
\resizebox{\columnwidth}{!}{%
\begin{tabular}{|c|c|c|c|c|}
\hline
 $C_{glob}$ & Controller &  $t^\mu_{lap}$ [s] & $\vert d^\mu \vert$ [m] & $\vert d^{max} \vert$ [m] \\ \hline \hline
                   &      \emph{Pure Pursuit}      & 8.74 & 0.115 & 0.33 \\ \cline{2-5} 
0.7                &      \gls{map} with linear      & \textbf{8.10} & 0.074 &  0.20 \\ \cline{2-5} 
                   &      \gls{map} with Pacejka     & 8.29 & \textbf{0.048} & \textbf{0.18} \\ \hline \hline
                   &     \emph{Pure Pursuit}       & N.C. (7.93*) & N.C. (0.220*) & N.C. (0.53*) \\ \cline{2-5} 
0.8                &     \gls{map} with linear       & N.C. (7.35**) &N.C.(0.55**)  & N.C.(0.32**) \\ \cline{2-5} 
                   &     \gls{map} with Pacejka      & \textbf{7.39} & \textbf{0.055} & \textbf{0.23} \\ \hline
\end{tabular}%
}
\caption{Lap times, average and maximum deviation from the trajectory for \emph{Pure Pursuit}, \gls{map} with linear tire model, and nominal \gls{map} with \emph{Pacejka} tire model. \\
N.C.: Not Completed\\ *: Crashed after one lap\\ **Unstable and crashed after two laps}
\label{tab:tire_res}
\end{table}

\subsection{Race Results}
The \gls{map} controller with \emph{Pacejka} tire parameters was deployed on the real racecar also in the \emph{2022 F1TENTH Autonomous Grand Prix Germany}\revdel{\footnote{https://ee.ethz.ch/news-and-events/d-itet-news-channel/2022/08/forzaeth-team-wins-gold.html \label{forzaeth}}}, obtaining the best performance in the field, winning the competition and achieving the fastest lap time, reaching speeds over \SI{8}{\meter/\second} and lateral accelerations of \SI{1.25}{g}. By measuring the maximal lateral force that could be applied to the racecar with a spring balance, resulting in \SI{45}{N} and the car's weight of \SI{3.5}{kg}, affirming the claim that the racecar can operate at the edge of friction.   

\section{CONCLUSIONS}
We presented \gls{map}, a highly performant racing controller suitable for computationally constrained hardware, that incorporates model dynamics while remaining simple to implement and operate. 
The proposed controller significantly outperforms \gls{sota} geometric controllers while conserving their benefits of design simplicity and real-time computational feasibility. 
%We show that by leveraging a tire model incorporating lateral tire slip increases tracking performance four-fold while scaling for higher velocities up to \SI{11}{m/s}, ultimately allowing for accurate trajectory tracking at high speed, by incorporating lateral tire slip. 
By leveraging model knowledge and incorporating tire models capturing lateral tire slip into the control law, tracking performance is increased four-fold during our tests at higher velocities of up to \SI{11}{m/s}.

Future work will focus on adaptive tire parameter estimation for the \gls{map} controller such that system identification would not be needed to be recomputed offline in different racing environments and an in-depth comparison with \gls{mpc} under realistic conditions. 
\rev{Furthermore, testing the behavior on a full-scaled racing car would be of great interest. The full \gls{map} code is available on Github: \url{https://github.com/ETH-PBL/MAP-Controller}.} 
%Moreover, we plan to exploit the same controller for other kind of vehicles and robots such as drones. 


%%%%%%%%%%%%%%%%%%%%%%%%%%%%%%%%%%%%%%%%%%%%%%%%%%%%%%%%%%%%%%%%%%%%%%%%%%%%%%%%

%%%%%%%%%%%%%%%%%%%%%%%%%%%%%%%%%%%%%%%%%%%%%%%%%%%%%%%%%%%%%%%%%%%%%%%%%%%%%%%%

%%%%%%%%%%%%%%%%%%%%%%%%%%%%%%%%%%%%%%%%%%%%%%%%%%%%%%%%%%%%%%%%%%%%%%%%%%%%%%%%

\section*{ACKNOWLEDGMENT}
The authors would like to thank all team members of the \emph{ForzaETH} \revdel{\textsuperscript{\ref{forzaeth}}} racing team, as well as Dr. Christian Vogt and Dr. Andrea Carron of ETH Zürich, for their constructive and fruitful algorithmic discussions.

%%%%%%%%%%%%%%%%%%%%%%%%%%%%%%%%%%%%%%%%%%%%%%%%%%%%%%%%%%%%%%%%%%%%%%%%%%%%%%%%

\bibliographystyle{IEEEtran}
\bibliography{main}

\end{document}